\documentclass[sigconf]{acmart}
\usepackage{multirow}
\usepackage{tabularx}
\usepackage{makecell}
\usepackage{float}
\usepackage{pifont} 
\newcommand{\cmark}{\ding{51}}%
\newcommand{\xmark}{\ding{55}}%
\AtBeginDocument{%
  }

\setcopyright{acmlicensed}
\copyrightyear{2025}
\acmYear{2025}
\acmDOI{XXXXXXX.XXXXXXX}
\acmConference[CODS '25]{Make sure to enter the correct
  conference title from your rights confirmation email}{Dec 17--20,
  2025}{Pune, India}
\acmISBN{978-1-4503-XXXX-X/2018/06}

\begin{document}

\title{Autograder+: A Multi-Faceted AI Framework for Rich Pedagogical Feedback in Programming Education}

\author{Vikrant Sahu}
\affiliation{%
  \institution{Indian Institute of Technology}
  \city{Bhilai}
  \state{Chhattisgarh}
  \country{India}
}
\email{vikrantsahu@iitbhilai.ac.in}

\author{Dr. Gagan Raj Gupta}
\affiliation{%
  \institution{Indian Institute of Technology}
  \city{Bhilai}
  \state{Chhattisgarh}
  \country{India}
}
\email{gagan@iitbhilai.ac.in}
\author{Raghav Borikar}
\affiliation{%
  \institution{Indian Institute of Technology}
  \city{Bhilai}
  \state{Chhattisgarh}
  \country{India}
}
\email{raghavbori@iitbhilai.ac.in}
\author{Nitin Mane}
\affiliation{%
  \institution{Indian Institute of Technology}
  \city{Bhilai}
  \state{Chhattisgarh}
  \country{India}
}
\email{nitingautam@iitbhilai.ac.in}

\renewcommand{\shortauthors}{Vikrant et al.}

\begin{abstract}

The rapid growth of programming education has outpaced traditional assessment tools, leaving faculty with limited means to provide meaningful, scalable feedback. Conventional autograders, while efficient, act as “black-box” systems that merely indicate pass/fail status, offering little instructional value or insight into the student’s approach. \textbf{Autograder+}\footnote{Code available at: \url{https://github.com/zvikrnt/Autograder-Plus}}, 
a comprehensive and intelligent framework designed to evolve autograding from a summative evaluation tool into a formative learning platform. addresses this gap through two unique features: (1) \textbf{feedback generation} via a fine-tuned Large Language Model (LLM), and (2) \textbf{visualization} of all student code submissions. The LLM is adapted via domain-specific fine-tuning on curated student code and expert annotations, ensuring feedback is pedagogically aligned and context-aware. In empirical evaluation across 600 student submissions across various programming problems, Autograder+ produced feedback with an average BERTScore F1 of 0.7658, demonstrating strong semantic alignment with expert-written feedback. To make visualizations meaningful, Autograder+ employs \textbf{contrastively learned embeddings} trained on 1,000 annotated submissions, which organize solutions into a performance-aware semantic space resulting in clusters of functionally similar appraoches. The framework also integrates a Prompt Pooling mechanism, allowing instructors to dynamically influence the LLM’s feedback style using a curated set of specialized prompts. By combining advanced AI feedback generation, semantic organization, and visualization, Autograder+ reduces evaluation workload while empowering educators to deliver targeted instruction and foster resilient learning outcomes.

\end{abstract}



\maketitle
\section{Introduction}
The escalating global demand for computational literacy has triggered a massive surge in enrollment in computer science courses at every level of education. This scaling presents an immense pedagogical challenge \cite{Messer_2024} providing timely, meaningful, and personalized feedback \cite{tang2024spherescalingpersonalizedfeedback} to an ever-growing body of students. The traditional method of meticulous manual code review by instructors or teaching assistants is logistically untenable in large-scale educational settings.

Automated assessment systems, commonly known as “autograders,” have emerged as a necessary solution to this scalability problem \cite{11058691}.  Platforms such as Gradescope \cite{10.1145/3051457.3051466} and Autolab automate \cite{li2022automatingcodereviewactivities} the execution and validation of student code, enabling immediate, objective, and scalable evaluation of functional correctness.  However, this efficiency comes at a significant pedagogical cost.  Most automated grading tools rely on dynamic test suites or static analysis and typically provide binary pass/fail outcomes or output comparisons, giving little insight into the student’s approach or conceptual errors.  Building comprehensive test suites is time‑consuming and incomplete test coverage can produce misleading feedback, failing to diagnose root causes or connect errors to underlying concepts .  Consequently, students often engage in trial-and-error loops—making surface-level changes to pass tests without resolving conceptual misunderstandings .  These tools, in essence, act as opaque arbiters of correctness, offering minimal educational scaffolding \cite{Messer_2024}. Beyond these limitations, automatic feedback systems also face challenges of trust: students frequently doubt the correctness of the system’s evaluations or the reliability of the feedback provided, which can diminish their confidence in the learning process. 

Recent work highlights the importance of richer feedback mechanisms. For instance, using context-aware LLMs with structured reasoning \cite{ Schneider2023TrustworthyAutograding} approaches—such as chain-of-thought prompting—can provide interpretable evaluations and actionable insights beyond mere correctness \cite{berthon2025languagebottleneckmodelsframework}. These approaches align more closely with pedagogical \cite{scholz2025partneringaipedagogicalfeedback} goals by diagnosing errors, guiding logic, and assisting conceptual understanding.

\begin{figure}[h]
\small
\centering
\includegraphics[width=0.95 \linewidth]{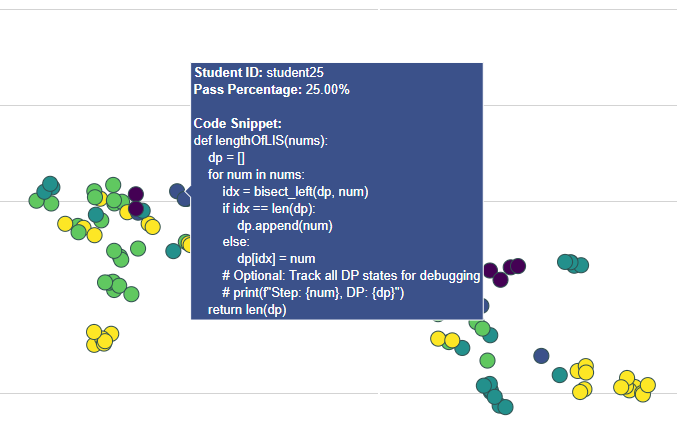}
\caption{A sample student code submission drawn from Interactive UMAP of code embeddings which are generated via the embedding model}
\label{fig:Figure 1}
\Description{}
\end{figure}

This paper contends that a fundamental disconnect exists between the assessment of functional correctness and the cultivation of conceptual understanding. To bridge this divide, we present Autograder+, a comprehensive and intelligent framework designed to evolve autograding from a summative evaluation tool into a formative learning \cite{wei2025conceptbasedrubricsimprovellm} platform. Secure, containerized code execution ensures robustness, while inference latency remains practical, averaging 11–13 seconds per response for selected models. Importantly, before any feedback is delivered to students, the generated output is validated by course instructors or TAs, ensuring that correct and non-hallucinated feedback reaches learners. Instructor-facing analytics further provide \textbf{actionable insights} into cohort-level trends, revealing performance patterns across assignments and time.The primary contributions of this work are as follows:
\begin{enumerate}
\item \textbf{A Complete Autograding Pipeline: }An end-to-end, modular framework combining secure sandboxed execution, static/ dynamic program analysis, and semantic evaluation for flexible, extensible code assessment.
\item \textbf{Innovative Feedback Enhancement: }A prompt pooling mechanism that dynamically injects expert-written prompts at inference, improving feedback quality.
\item \textbf{Concept-Aware Instructor Analytics:} Interactive UMAP visualizations of code embeddings (Fig. \ref{fig:Figure 1}) learned via contrastive fine-tuning, reveal common strategies, misconceptions, and outliers for targeted pedagogical correction.
\end{enumerate}

\textbf{Autograder+} accomplishes this by augmenting traditional program validation with deep semantic analysis, sophisticated AI-driven feedback, and interactive visual analytics for instructors \cite{Bassner_2024}. In empirical evaluation on 600 student submissions, Autograder+ achieved an average BERTScore F1 of 0.75 against TA- written gold-standard feedback, demonstrating that its generated explanations closely align with human instructors.

\section{Related Work}
The design and philosophy of Autograder+ are built upon a rich history of research spanning automated assessment \cite{hicke2023aitaintelligentquestionanswerteaching}, program analysis, educational data mining, and the revolutionary advancements in artificial intelligence for code \cite{software1010002}.

\subsection{Traditional Autograding Systems}
The foundation of automated assessment in programming was laid by systems prioritizing scalability and objective evaluation\cite{software1010002}.  Seminal platforms like Autolab and Gradescope defined the paradigm of test‑case‑driven assessment, executing student code in a sandbox and comparing outputs against expected results\cite{AutomatedGradingReview2023}.  Their impact has been transformative, enabling instructors to manage assignments in massive open online courses (MOOCs) and large university classes, but the pedagogical model is inherently limited\cite{AutomatedGradingReview2023}.  The “black‑box” nature of this testing provides little insight into the student’s cognitive process or algorithmic strategy; feedback typically consists of binary pass/fail signals or cryptic output diffs\cite{AutomatedGradingReview2023}.

\subsection{Program Analysis for Educational Feedback}
Recognizing the limitations of simple input/output testing, researchers have long sought to “open the black box” using techniques from program analysis. \textbf{Static analysis} tools inspect the source code without running it, typically by constructing an Abstract Syntax Tree (AST) \cite{abdelmalak2025astguidedllmapproachsvrf} or a control-flow graph \cite{li2025sclaautomatedsmartcontract}. This allows for detection of syntax errors, violations of coding style, and structural anti-patterns, enabling systems to provide students with feedback on the form and structure of their code\cite{software1010002}. \textbf{Dynamic analysis} tools execute the code to observe its runtime behaviour, catching exceptions, logical errors, and performance issues that static analysis might miss.  While these methods offer more granular feedback than traditional autograders, they often focus on technical aspects rather than student intent, and typically fail to diagnose higher-order conceptual errors\cite{AutomatedGradingReview2023}.  Moreover, such feedback is rarely mapped explicitly to curricular learning objectives\cite{9962650}.

\subsection{Large Language Models and Prompt Engineering in Education}
The advent of large language models (LLMs) pre-trained on vast datasets of code, such as CodeBERT \cite{feng2020codebertpretrainedmodelprogramming}, CodeT5+ \cite{wang2023codet5opencodelarge}, and the Qwen series \cite{qwen2025qwen25technicalreport}, has unlocked new possibilities for automated pedagogical support \cite{10825949}. These models exhibit a profound ability to comprehend, summarize, and generate code \cite{akyash2025stepgradegradingprogrammingassignments}. Initial educational applications focused on using these models as explainers’’ or translators’’. More recently, a wave of research has explored their potential for generating formative feedback on student programming assignments \cite{Pathak_2025}. These studies confirm that LLMs can produce fluent, human-like feedback that transcends syntax. However, many of these efforts treat the LLM as a standalone component, disconnected from a robust execution pipeline \cite{10673924}. Furthermore, controlling the output of these powerful models to be pedagogically sound is a significant challenge. This has given rise to the field of \textbf{prompt engineering} \cite{sahoo2025systematicsurveypromptengineering}, where the input given to the model is carefully crafted to steer its output. Our work on Dynamic Prompt Pooling builds directly on this idea, but automates the selection of the steering prompt based on semantic analysis. While others have used LLMs for feedback, Autograder+ distinguishes itself by embedding two distinct, advanced AI modeling strategies within a complete autograding framework, ensuring feedback is grounded in the code’s actual runtime behavior and enhanced by dynamic, pedagogically-informed prompt steering \cite{11025882}.

\section{AI-Driven Semantic Feedback Models}
We now elaborate on the two primary AI model variants that power Autograder+. These models represent distinct strategies for generating high-quality pedagogical feedback. The first is a direct approach focused on fine-tuning a generative model, while the second employs a more sophisticated method of structuring the semantic space through contrastive learning \cite{10825949}.

\subsection{Model Variant 1: Fine-Tuned LLM}
The first variant represents a direct and powerful approach to feedback generation. It involves taking a pre-trained Large Language Model (LLM), and specializing it through supervised fine-tuning. We leverage a subset of the \verb|nvidia/openreasoningcode| dataset \cite{ahmad2025opencodereasoning} consisting of \verb|problem & code| pairs. The model is trained to generate the \verb|debugging| \verb|insight| when conditioned on the \verb|student_code|. By learning from examples, the model learns to identify common errors and articulate explanations in a manner that is both conceptually precise and accessible to students. While highly effective, this approach treats the code and its performance as input text, without explicitly structuring the underlying semantic relationships \cite{Roussinov2023} between different student solutions.
\subsection{Model Variant 2: Contrastively Fine-Tuned Embedding Model}
The second, structurally distinct variant enhances instructor-facing analytics. This approach fine-tunes the embedding model itself to create a performance-aware semantic space where the geometric arrangement of code embeddings reflects their functional correctness. These structured embeddings are then used to drive UMAP visualizations \cite{mcinnes2020umapuniformmanifoldapproximation}, enabling instructors to identify clusters of correct, partially correct, and incorrect solutions, as well as recurring misconceptions or outlier strategies. This is achieved by fine-tuning a base code embedding model using a combination of powerful contrastive loss functions \cite{lee2025similaritiesembeddingscontrastivelearning}.
\subsubsection{Multi-Label Supervised Contrastive Finetuning: }
To create embeddings that are aware of both the problem type (e.g., Fibonacci, Palindrome) and the correctness of the solution (e.g., \verb|PASS|, \verb|PARTIAL|, \verb|FAIL|), we fine-tune a base code embedding model using a \textbf{Multi-Label Supervised Contrastive Loss (MulSupCon)}, inspired by the work of Zhang et al. \cite{zhang2024multi}. This approach teaches the model to group similar solutions in the embedding space based on shared characteristics defined by their labels.

\textbf{Mathematical Formulation: }Let $\mathcal{B} = \{(z_i, y_i)\}_{i=1}^N$
be a batch of \(N\) samples, where \(z_i \in \mathbb{R}^D\) is the \(D\)-dimensional embedding of a code snippet and \(y_i \in \{0,1\}^C\) is its corresponding multi‐hot label vector over \(C\) classes (e.g., \verb|problem_q6|, \verb|tier_PASS|).

First, we compute the pairwise cosine similarity \cite{7577578} between all
normalized embeddings in the batch. This forms a similarity matrix
$\mathbf{S} \in \mathbb{R}^{N \times N}$, where $\mathbf{S}_{ij} =
\mathbf{z}_i \cdot \mathbf{z}_j$.

These similarities are then scaled by a temperature parameter $\tau >
0$ to produce logits, which control the sharpness of the probability
distribution.

\[
\text{logits}_{ij} = \frac{\mathbf{S}_{ij}}{\tau} = \frac{\mathbf{z}_i
\cdot \mathbf{z}_j}{\tau}
\]

The log-probability \cite{soru2025leveraginglogprobabilitieslanguage} of correctly identifying a sample $j$ as a
positive for an anchor sample $i$ (among all other samples $m \neq i$
in the batch) is given by the log-softmax function:

\[
\log P_{ij} = \text{logits}_{ij} - \log \sum_{m=1, m \neq i}^{N}
\exp(\text{logits}_{im})
\]
For an anchor sample $i$ and a specific class $k$, the set of indices
of its positive samples, $P(i, k)$, includes all other samples $j$ in
the batch that also possess class $k$.

\[
P(i, k) = \{ j \in \{1, \dots, N\} \setminus \{i\} \mid
\mathbf{y}_{ik} = 1 \text{ and } \mathbf{y}_{jk} = 1 \}
\]

The supervised contrastive loss for anchor $i$ *with respect to class
k* is the average of the negative log-probabilities over all its
positive samples for that class \cite{khosla2020supervised}. This loss is only calculated if the
positive set $P(i, k)$ is not empty ($|P(i, k)| > 0$).

\[
\mathcal{L}_{i,k} = \begin{cases}
-\frac{1}{|P(i, k)|} \sum_{j \in P(i, k)} \log P_{ij} & \text{if }
|P(i, k)| > 0 \\
0 & \text{otherwise}
\end{cases}
\]

The total loss for a single anchor sample $i$ is the sum of its
per-class losses, calculated only for the classes it actually
possesses. The multi-hot label vector $\mathbf{y}_i$ acts as a mask
for this summation.

\[
\mathcal{L}_i = \sum_{k=1}^{C} \mathbf{y}_{ik} \cdot \mathcal{L}_{i,k}
\]

Finally, the total Multi-Label Supervised Contrastive Loss for the
entire batch $\mathcal{B}$ is the mean of the individual anchor
losses. We only average over anchors that have at least one label to
avoid division by zero if a sample has no labels. Let $\mathcal{I}^+ =
\{i \mid \sum_{k=1}^C \mathbf{y}_{ik} > 0\}$ be the set of indices of
anchors with at least one label.

\[
\mathcal{L}_{\text{MulSupCon}} = \frac{1}{|\mathcal{I}^+|} \sum_{i \in
\mathcal{I}^+} \mathcal{L}_i
\]
\subsubsection{Multiple Negatives Ranking (MNR) Loss: }
We further refine the model's ability to distinguish between samples with a complementary training objective by incorporating the \textbf{Multiple Negatives Ranking (MNR) Loss} \cite{inproceedingsmnrloss}, which is highly effective for retrieval-oriented tasks. For a given positive pair (anchor, positive), MNR loss treats all other samples in the batch as hard negatives and uses a standard cross-entropy loss to train the model to assign a higher similarity score to the positive pair than to any of the negative pairs. This directly optimizes a ranking objective. \\

We combine these two losses using a weighting factor $\alpha$:
\begin{equation}
\mathcal{L}_{\text{Total}} = \alpha \cdot \mathcal{L}_{\text{MulSupCon}} + (1 - \alpha) \cdot \mathcal{L}_{\text{MNR}}
\end{equation}
By training our embedding model with this hybrid loss function, the resulting semantic space becomes highly structured along multiple axes simultaneously. This performance-aware embedding is a far more potent input for the instructor analytics engine.\cite{zhang2024multi}

\begin{figure*}[h]
\small
  \centering
  \includegraphics[width=\linewidth]{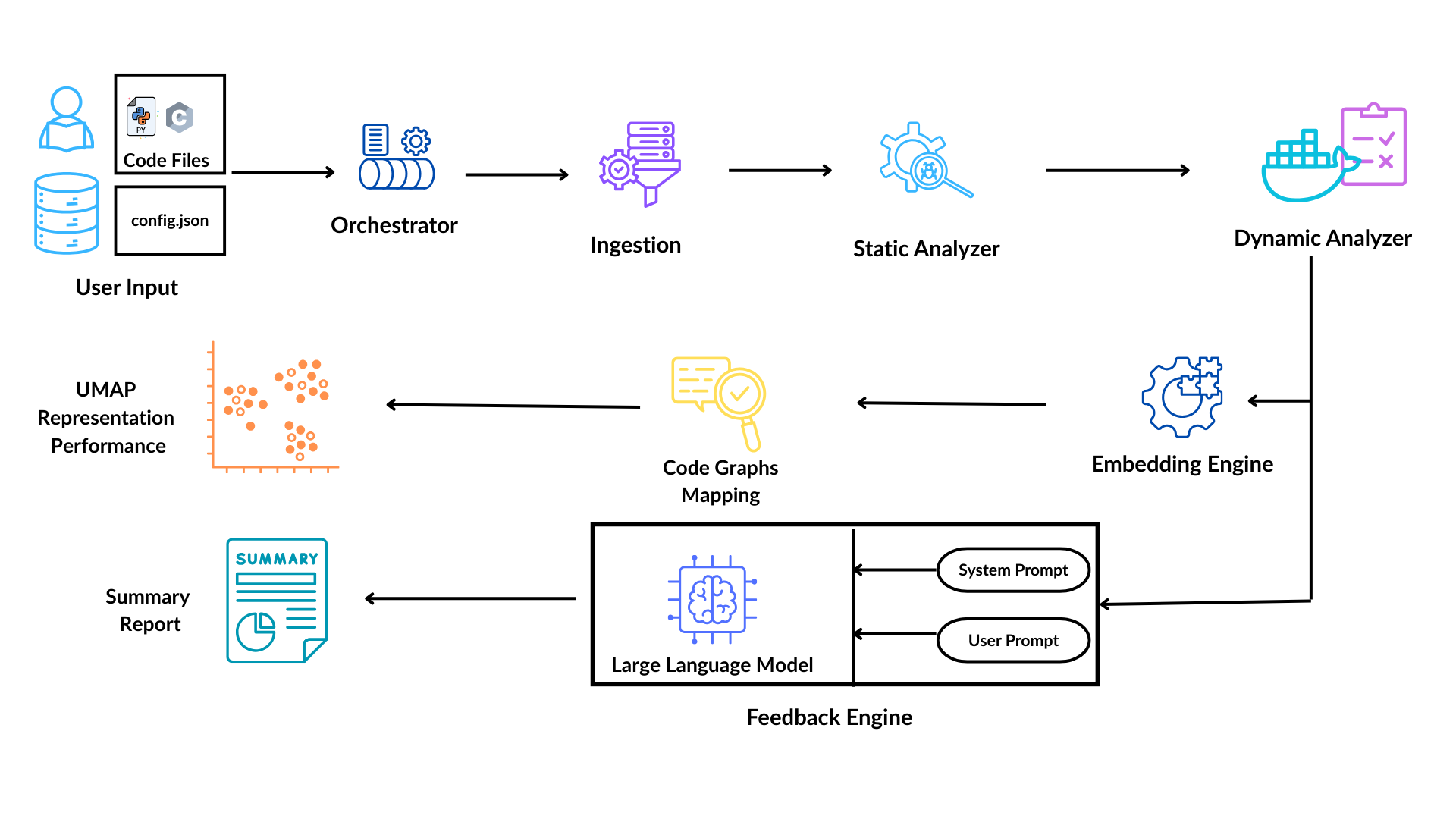}
  \caption{End‐to‐end architecture of Autograder+.}
  \Description{}
\end{figure*}

\section{The Autograder+ Framework}
Autograder+ is architected as a modular, multi-stage pipeline designed to systematically process student submissions, enriching them at each step with layers of analysis. This design ensures that the final feedback is a holistic synthesis of functional correctness, structural integrity, and deep semantic understanding \cite{10825949}.

\subsection{System Architecture}
The journey of a student's code through the Autograder+ framework is a structured progression, orchestrated by a series of specialized engines. The core components and data flow are as follows:\\
\begin{enumerate}
    \item \textbf{Code Ingestion: }The process begins with the \verb|Ingestor| module. It is designed with the flexibility to handle common submission formats. The ingestor reads the source code, links it with the corresponding assignment configuration file—a JSON file specifying the problem description, test cases, execution parameters, and language—and encapsulates this information into a standardized data object that flows through the rest of the pipeline.

    \item \textbf{Static Analysis Engine: }The submission then passes through the \verb|Static Analyzer|. It performs a fast, low-cost analysis of the code without executing it. By parsing the code into an Abstract Syntax Tree (AST), it can efficiently validate syntax, count key structural elements (e.g., number of loops, function definitions), verify adherence to assignment constraints (e.g., presence of a required function name/absence of forbidden libraries), and flag basic anti-patterns \cite{11025882}. This stage provides an immediate structural and syntactic health check .
    \item \textbf{Dynamic Execution Engine: }Code that is structurally sound proceeds to the \verb|Dynamic Analyzer|, the crucible to test functional correctness. To ensure safety, security, and reproducibility, this engine leverages Docker containers. Each test case for a submission is executed in a fresh, isolated container, effectively sandboxing the code to prevent filesystem contamination or network access and to enforce resource limits (CPU, memory). This one-shot container strategy guarantees that each test run is independent and clean. The engine meticulously captures the program's standard output (stdout), standard error (stderr), and exit code, comparing them against the expected outcomes defined in the assignment configuration. This stage delivers the definitive ground truth about the code's runtime behavior \cite{10825949}.

    \item \textbf{The Semantic Core: }Following dynamic analysis, the complete submission package—source code, static analysis results, and the detailed execution trace—is passed to the semantic core. It is here that an LLM is employed to generate feedback. The core contains two key sub-modules:\\
\verb|Embedding Engine|: This module uses an embedding model to convert the student's source code into a high-dimensional vector embedding. This embedding captures the code's semantic meaning, abstracting away from surface-level syntax to represent its deeper algorithmic intent. \cite{zhang2024multi}.\\
\verb|Feedback| \verb|Engine|: This module orchestrates the generation of pedagogical feedback. It takes the code and the dynamic analysis results. Crucially, it houses the Prompt Pooling mechanism, which enhances the final output. The engine then calls the generative model (Base or Fine Tuned) to produce the final textual feedback. \cite{Banihashem2025HybridFeedback}.

    \item \textbf{Reporting and Analytics:}\\
\verb|Feedback Generator|: This module aggregates all the structured information gathered throughout the pipeline and compiles comprehensive reports. It generates an individual Markdown report for each student, presenting the static analysis, a test-by-test breakdown of dynamic results, and the rich, qualitative AI-generated feedback. It also creates aggregated summaries, such as a class-wide CSV file for grade-keeping and high-level review \cite{ai6020035}.\\
\verb|Analytics Engine|: This final engine serves the instructor. It collects the semantic code embeddings from every submission in the class and uses the Uniform Manifold Approximation and Projection (UMAP) algorithm to project them into an interactive 2D scatter plot. This visualization provides an intuitive map of the class's collective problem-solving approaches, transforming raw submission data into actionable pedagogical insights. \cite{frederick-eneye-etal-2025-advances}
\end{enumerate}

\subsection{Feedback Enhancement via Prompt Pooling}
A key innovation within the Autograder+ framework is the Prompt Pooling mechanism, which enhances the pedagogical quality of feedback from any underlying generative model. This technique provides a lightweight and dynamic method for steering the LLM's focus at inference time, ensuring that its output is not only technically accurate but also aligned with a specific, contextually relevant instructional goal \cite{ Chan2024AIFeedbackOutcomes}.
The mechanism operates as follows:
\begin{figure}[h]
\small
\centering
\includegraphics[width=\linewidth]{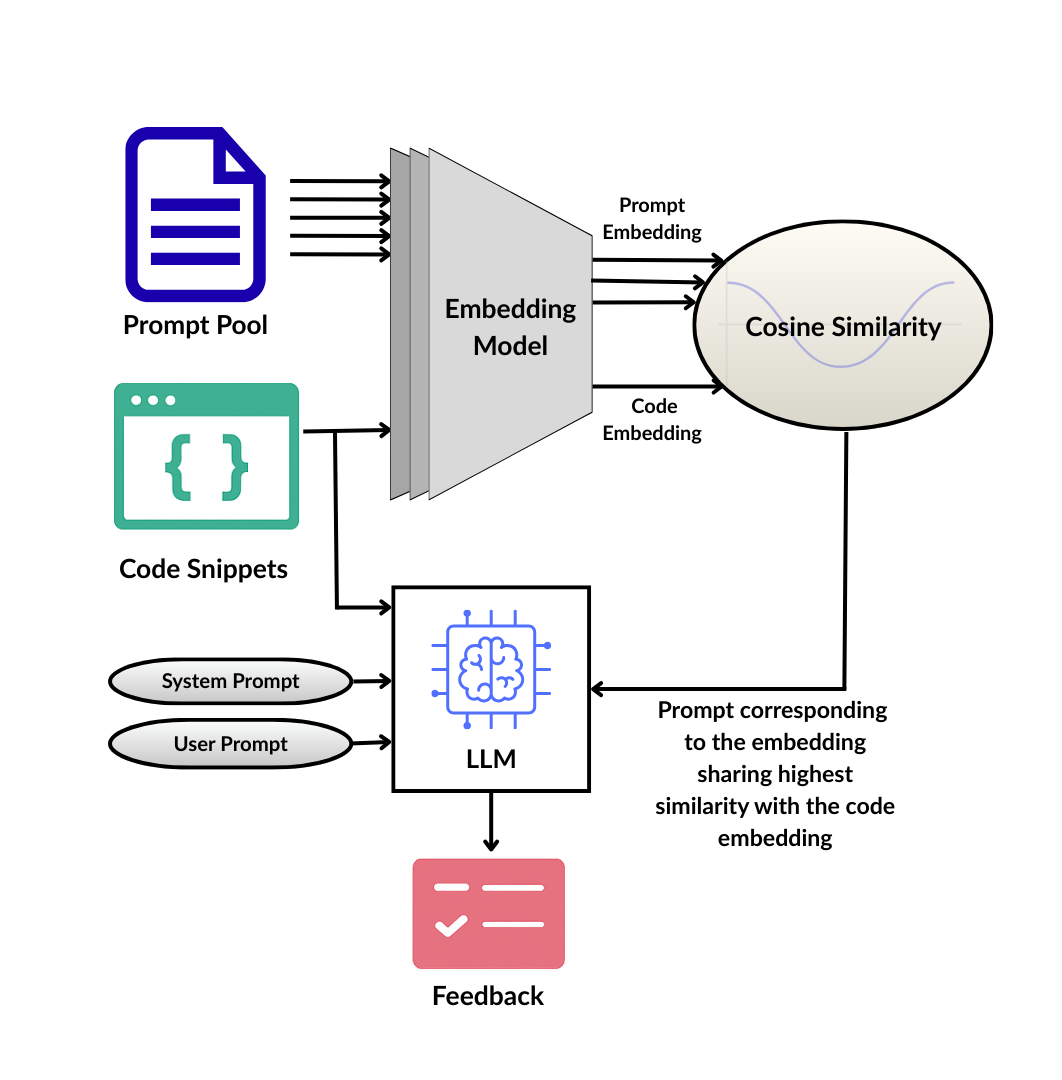}
\caption{Architecture of the prompt‐pooling framework}
\Description{Selection of a prompt to be passed to the LLM takes place based on the magnitude of cosine similarity between the corresponding embedding and the passed code embedding.}
\end{figure}

\begin{enumerate}
\item \textbf{Curate a Prompt Pool:} A repository of expert-written instructional prompts is created by instructors or curriculum designers. Each prompt is designed to focus an LLM's analysis on a specific programming concept (e.g., recursion base cases, loop termination conditions), error type (e.g., IndexError, TypeError), or pedagogical strategy \cite{acm2025feedbackFormative}.
\item \textbf{Pre-computation of Prompt Embeddings:} At initialization, the framework uses an external embedding generator (e.g., a Sentence-BERT model) to compute a high-dimensional vector embedding for every prompt in the pool. These prompt embeddings are then cached for efficient retrieval \cite{tseng2025codev}.
\item \textbf{Runtime Code Embedding:} When a student submission is processed, the \textit{same} embedding generator is used to create a vector embedding for the student's code snippet .
\item \textbf{Semantic Similarity Search:} The system then calculates the cosine similarity between the student's code embedding and every cached prompt embedding in the pool. This identifies which instructional prompt is most semantically relevant to the student's specific solution.
\item \textbf{Dynamic Prompt Injection:} The prompt corresponding to the highest cosine similarity score is selected as the ``best-matching'' instructional focus. This selected prompt is then appended to the context provided to the generative LLM. The final request sent to the LLM includes the standard system prompt defining the AI's persona \cite{Idowu2024DebiasingAI}, the user prompt containing the student's code and error logs (if any), and this dynamically chosen instructional prompt.
\end{enumerate}
This approach provides remarkable flexibility. An instructor can easily refine the system's pedagogical behavior by simply adding or editing the text-based prompts in the pool, allowing the framework to be adapted to different course levels or topics with minimal technical overhead \cite{Banihashem2025HybridFeedback}.

\begin{table*}[h]
\small
    \centering
    \caption{A sample I/O from the framework}
    \label{tab: Table-1}
    \begin{tabular}{llcll}
        \toprule
        Question & Code Snippet & Test Cases Passed & Prompt Selected & Feedback\\
        \midrule
        \makecell[l]{Implement a function to\\compute the nth Fibonacci\\number. The Fibonacci sequence\\is defined as follows:\\F(0) = 0, F(1) = 1, and \\F(n) = F(n-1) + F(n-2) for n > 1.\\The function should handle both\\small and large inputs efficiently.}
 & \makecell[l]{def fibonacci(n):\\a, b = 0, 1\\for \_ in range(n):\\c = a + b\\a, b = b, a + b\\return a} & 6/7 & \makecell[l]{Advice: Pay attention\\to the loop constructs\\ Check for correct\\initialization, termination\\conditions,and potential\\off-by-one or infinite\\loop errors} & \makecell[l]{<output> The issue lies in the\\line `a, b = b, a + b`,where\\you're incorrectly updating the\\values of `a` and `b`. It should be\\`a, b = c, a + b` to\\correctly calculate the\\Fibonacci sequence. </output>}\\
        \midrule
         \makecell[l]{Given an integer array nums,\\return the length of the longest\\strictly increasing subsequence.\\Input: nums = [10,9,2,5,3,7,101,18]\\Output: 4 Explanation: The longest\\ increasing subsequence is [2,3,7,101],\\therefore the length is 4.} & \makecell[l]{def lengthOfLIS(nums):\\if not nums: return 0\\dp = [0] * len(nums)\\for i in range(len(nums)):\\for j in range(i):\\
            if nums[i] > nums[j]:\\
                dp[i] = max(dp[i], dp[j] + 1)\\
    return max(dp)} &0/4& \makecell[l]{Advice: Pay attention\\to dynamic programming\\concepts. Look for\\evidence of memoization\\or tabulation to avoid\\re-computing results}& \makecell[l]{<output>It seems like you're\\comparing elements correctly\\but the issue might be with\\how you're handling the DP\\array. Are you initializing it\\properly and updating its values\\accordingly?</output>}\\
        \bottomrule
    \end{tabular}
\end{table*}

A representative example of this process is shown in Table 1, where a student’s code is analyzed. The framework not only executes the code but also selects a semantically relevant instructional prompt (“loop constructs”) and generates feedback that pinpoints the precise logical error. This example demonstrates how Autograder+ moves beyond binary correctness, producing feedback that is both actionable and conceptually targeted.

\section{Experiments and Results}
To rigorously evaluate the Autograder+ framework and compare the efficacy of our two primary AI model variants, a comprehensive experimental study was conducted.

\subsection{Datasets}
To train, validate, and evaluate the proposed AI models, we utilized a combination of internally collected student submissions and a large-scale external code corpus. These datasets serve distinct purposes, as described below.

\begin{table}[h]
\small
    \centering
    \caption{Datasets Collected/Used }
    \label{tab: Table 2}
    \begin{tabular}{cccc} 
        \toprule
        Dataset&Source&Size&Purpose\\
        \midrule
        \makecell{NVidia\\ORC}&\makecell{External\\Corpus}&\makecell{171,000 Code\\Examples}&\makecell{Preliminary\\Fine Tuning of\\Base Model}\\
        \midrule
        IITBh PW25S &\makecell{Programming\\ Workshop}&1000 Submissions&\makecell{Contrastive\\Fine-Tuning\\ of Embedding Model}\\
        \midrule
        IITBh PC25&\makecell{Internal\\Programming\\Course} &600 Submissions&Evaluation\\
        \bottomrule
    \end{tabular}
\end{table}

\subsubsection{External Corpus for Foundational Fine-Tuning}
To enhance the general code reasoning capabilities of our base generative model we performed fine-tuning using a subset of \texttt{nvidia/openreasoning} \texttt{code} dataset \cite{ahmad2025opencodereasoning}, which was further augmented and contained $\sim$ 15,000 code examples. 

\subsubsection{Institute-Collected Student Submissions}
We curated two distinct datasets from our academic and workshop activities, each tailored to a specific modeling approach. \cite{Green2025}

\textbf{IITBh PW25S: }1,000 student submissions collected during a programming workshop at IIT Bhilai. Each submission was automatically labeled by our Dynamic Analyzer as \texttt{pass, partial-pass, or fail} based on test-case results.\cite{Andersen2025Fairness}.These labels enabled construction of \texttt{[anchor, positive, negative]} triplets for contrastive training of the embedding model, creating a performance-aware semantic space, where positive samples are drawn from the same correctness class as the anchor, and negative samples are drawn from a different class. This process enables the model to learn a performance-aware semantic space without the need for manual feedback annotation.

\textbf{IITBh PC25: }600 student submissions collected from internal programming courses. \cite{yousef2025begrading} spanning 20 LeetCode-style algorithmic problems(e.g., Fibonacci, Disarium). \cite{estevez2024evaluatingLLMFeedback} This dataset is particularly valuable as it contains a diverse range of correct solutions, partially correct attempts, and common logical and syntax errors. \cite{Henderson2025AIvsTeacherFeedback} Each of them was manually evaluated by Teaching Assistants (TAs), who provided "gold-standard" pedagogical feedback. \cite{Er2025InstructorVsAI} This collection of \verb|(submission, TA_Feedback)| pairs served as the primary \textbf{validation corpus} for our Feedback models while the expert TA feedback serves as the \textbf{reference ground truth} for SBERT and BERTScore based evaluation. 

\subsection{Evaluation Metrics}
The quality of the AI-generated feedback will be quantified using two key metrics that measure semantic alignment with the human-written reference feedback.\ \cite{DAI2024100299}\\
\textbf{SBERT Cosine Similarity: }This metric evaluates the semantic similarity at the sentence level. It measures the cosine of the angle between the sentence embeddings of the generated feedback and the reference feedback.\\
\textbf{BERTScores: }These metrics operate at the token level, computing a similarity score for each token in the generated feedback against tokens in the reference feedback. It provides more granular precision, recall, and F1-scores, capturing lexical overlap in a context-aware manner.\ \cite{silva2025assessingllmprogramming}

\subsection{Main Results}
We first evaluated several state-of-the-art large language models (LLMs) integrated directly into the Autograder+ feedback pipeline without any domain-specific fine-tuning or prompt pooling. The aim was to establish a realistic performance baseline against which subsequent enhancements could be measured.\cite{Schneider2023TrustworthyAutograding}

Table 3 summarizes these results. Four key metrics are reported: \textbf{BERTScore F1}, \textbf{Precision}, \textbf{Recall}, and \textbf{SBERT Cosine Similarity}—each measuring how closely the AI-generated feedback aligns semantically with gold-standard, TA-written feedback. Higher values indicate closer semantic alignment with expert feedback.\ \cite{federiakin2024promptSkill}

Among the tested models, \textbf{falcon3:10b} \cite{Falcon3} (0.7435) and \underline{llama3.2:3b} \cite{Llama_3_2_3B_HuggingFace} (0.7235) achieve the highest BERTScore F1 values, indicating stronger semantic alignment with human feedback \cite{estevez2024evaluatingLLMFeedback}, while other models such as qwen3:8b and phi4-reasoning lag significantly in both precision and recall \cite{acm2025feedbackFormative}. The SBERT cosine similarity scores, although lower in absolute terms, follow the same trend, with \underline{falcon3:10b} and \textbf{llama3.2:3b} leading the chart and inherently providing a contextually relevant feedback even before fine-tuning. \cite{10.1145/3634814.3634816}.

\begin{table}[h]
\small
    \centering
    \caption{Baseline Results: The Results present BERT Scores for the models that were incorporated directly into the framework. (w/o fine tuning or prompt pooling)  }
    \label{tab: Table-3}
    \begin{tabular}{ccccc} 
        \toprule
        Model & \makecell{Avg.\\F1} &\makecell{Avg.\\Precision}& \makecell{Avg.\\Recall}&\makecell{Avg. Cosine\\Similarity}\\
        \midrule
        qwen3:8b&0.3367&0.3253&0.3492&0.1854\\
        {deepseek-coder:33b}&{0.7212}&{0.7020}&\textbf{0.7421}&{0.3241}\\
\textbf{falcon3:10b}&\textbf{0.7435}&\textbf{0.7485}&{0.7390}&\underline{0.3449}\\
        \underline{llama3.2:3b}&\underline{0.7235}&\underline{0.7072}&\underline{0.7412}&\textbf{0.3667}\\
        phi4-reasoning&0.3211&0.3144&0.3289&0.1100\\
        \bottomrule
    \end{tabular}
\end{table}
To further refine our selection for subsequent experiments, we evaluated the average inference time of each baseline model when integrated into the Autograder+ pipeline. This metric reflects both the computational cost and the practical feasibility of deploying these models in classroom settings, where real-time or near real-time feedback is crucial.

Table 4 summarizes the inference latency (in seconds per response) observed during baseline evaluation. As expected, reasoning based models such as \texttt{phi4-reasoning} \cite{microsoft-phi-4-reasoning-huggingface} incur significantly higher latency, making them less suitable for scalable deployment while \texttt{deepseek-coder:33b} \cite{guo2024deepseekcoderlargelanguagemodel} is  moderately efficient and comparitively less aligned with human feedback. 
\begin{table}[h]
\small
    \centering
    \caption{Inference Time Analysis: Average time per response measured in seconds across the baseline models.}
    \label{tab: Table-4}
    \begin{tabular}{cc}
        \toprule
        Model & \makecell{Avg. Inference Time\\(seconds/response)} \\
        \midrule
        qwen3:8 & 63s \\
        deepseek-coder:33b & 20.6s \\
        \underline{falcon3:10b} & \underline{13.2s} \\
        \textbf{llama3.2:3b} & \textbf{11.8s} \\
        phi4-reasoning & 60.3 \\
        \bottomrule
    \end{tabular}
\end{table}
Balancing both semantic quality and computational feasibility, we selected \textbf{llama3.2:3b} and \underline{falcon3:10b} as candidates for the next stage of our study.

Having established the baseline, we integrated fine-tuning, prompt pooling, and their combination into the Autograder+ framework. Table 5 presents results from these enhanced configurations \cite{Andersen2025Fairness}.

Table 5 summarizes the impact of adding question text, prompt pooling, and fine-tuning to our Autograder+ framework. The most consistent improvements come from prompt pooling, which raises both lexical (BERT F1/Precision) and semantic (SBERT cosine) scores across models, with \textbf{falcon3-10B (Base)} achieving the best lexical match (F1 = 0.7658, Precision = 0.7706) and \textbf{llama3.2-3B (Base)} reaching the highest semantic similarity (SBERT = 0.3924). Including the question text yields smaller, mixed gains, suggesting model sensitivity to input format. Interestingly, fine-tuning did not outperform the strong base models, and in most cases resulted in a slightly reduced performance. We attribute this to the nature of our fine-tuning data, which was augmented and partly synthetic, introducing noise and stylistic artifacts that likely led to overfitting and reduced generalization. Additionally, distributional mismatch between the training prompts and evaluation setup, as well as potential overspecialization during fine-tuning, may have contributed to the observed drops. Overall, prompt pooling emerges as the most robust enhancement, while fine-tuning provides limited benefit under our current (augmented) dataset.

It is also important to note that the comparison involves models of different sizes (3B vs. 10B parameters). While falcon3-10B naturally benefits from its larger capacity, our results show that llama3.2-3B, despite being smaller, can achieve competitive or even superior semantic similarity when combined with prompt pooling. This highlights that architectural choices and configuration strategies can sometimes outweigh raw model scale in the context of automatic grading.
\begin{table*}[t]
    \centering
    \caption{Results across various configurations for Base and Fine Tuned Models}
    \label{tab: Table 5}
    \begin{tabular}{ccccccc}
        \toprule
        Model & Type & Question & Prompt Pool & Avg. BERT F1 & \makecell{Avg. BERT\\Precision} & \makecell{Avg. SBERT\\Cosine}\\
        \midrule
        \multirow{6}{*}{falcon3:10b} 
        & \multirow{3}{*}{Base} & \xmark & \xmark & 0.7361 & {0.7372}& {0.3307} \\
        &  & \cmark & \xmark & 0.7435&0.7485&0.3449 \\
        &  & \cmark & \cmark & \textbf{0.7658} & \textbf{0.7706} & \textbf{0.3725} \\
        \cline{2-7}
        & \multirow{3}{*}{FT}   & \xmark & \xmark & 0.7128 & 0.6812 & 0.3298 \\
        & & \cmark & \xmark & 0.7092 & 0.6714 & 0.3317 \\
        & & \cmark & \cmark & \textbf{0.7340} & \textbf{0.7286} & \textbf{0.3459} \\
        \midrule
        \multirow{6}{*}{llama3.2:3b} 
        & \multirow{3}{*}{Base} & \xmark & \xmark & 0.7134 & 0.6951 & 0.3506 \\
        & & \cmark & \xmark & 0.7235 & 0.7072 & 0.3667 \\
        & & \cmark & \cmark & \textbf{0.7452} & \textbf{0.7315} & \textbf{0.3924} \\
        \cline{2-7}
        & \multirow{3}{*}{FT} & \xmark & \xmark & 0.7056 & 0.6851 & 0.3321 \\
        & & \cmark & \xmark & 0.7182 & 0.6954 & 0.3537 \\
        & & \cmark & \cmark & \textbf{0.7369} & \textbf{0.7321} & \textbf{0.3788} \\
        \bottomrule
    \end{tabular}
\end{table*}

Beyond numerical evaluation, understanding how students approach problems and where misconceptions cluster is critical for targeted instruction \cite{frederick-eneye-etal-2025-advances}.To complement the quantitative metrics in Tables 3 and 5, we leverage the performance-aware embeddings generated by Autograder+ to visualize entire cohorts’ solution spaces. Using interactive UMAP projections, code embeddings reveal distinct clusters of correct, partially correct, and incorrect solutions when attempted using diverse approaches, enabling instructors to spot recurring error patterns, strategies used, and isolated outlier cases at a glance. The following section presents these qualitative analytics, illustrating how Autograder+ transforms raw performance data into actionable pedagogical insight \cite{Xu2025AIFeedbackMotivation}.
\begin{figure}[h]
\centering
\includegraphics[width=\linewidth]{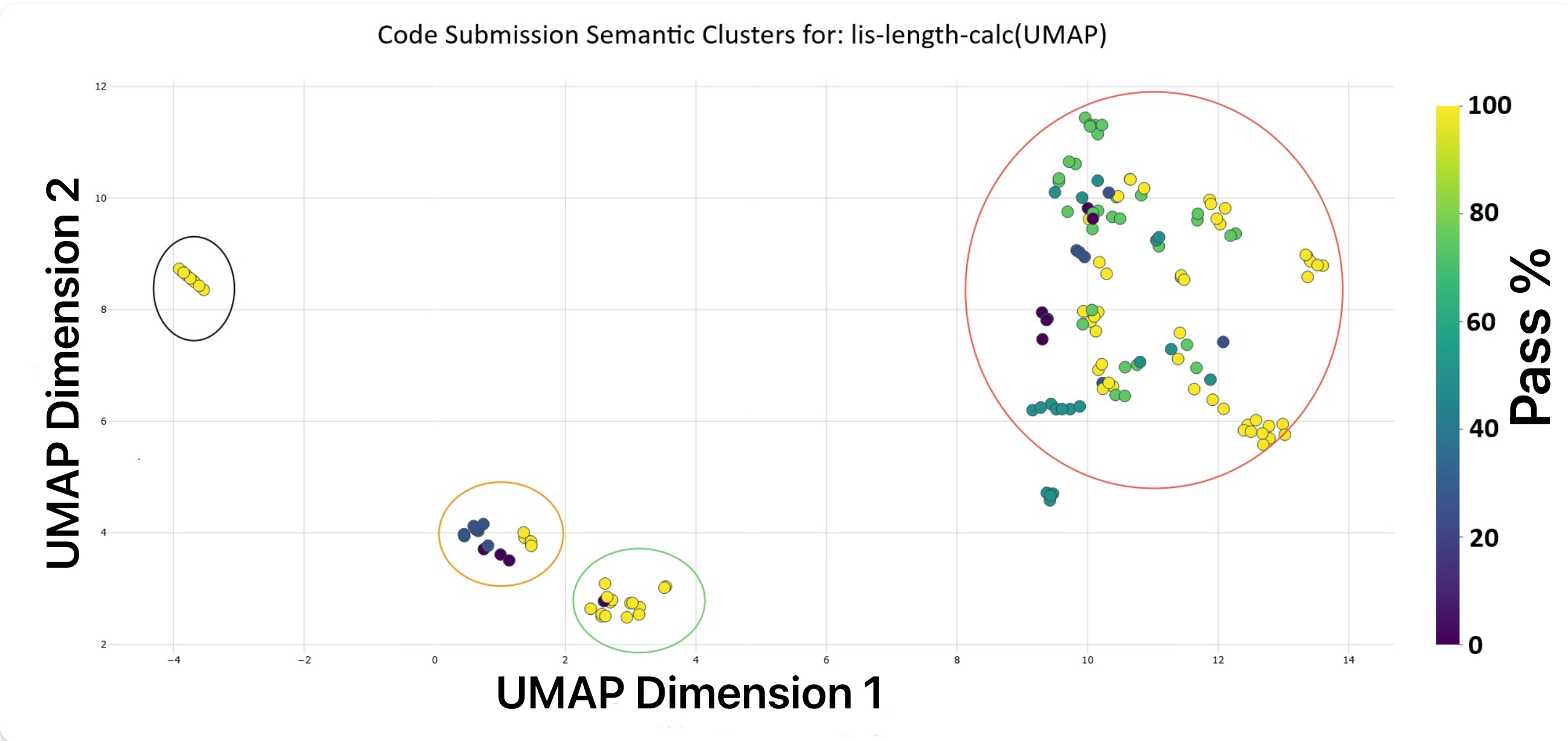}
\caption{UMAP projection of embeddings of code as generated by the embedding model before contrastive fine tuning. Each point is a single submission, shaded by its performance (e.g., Light=PASS, Dim=PARTIAL PASS, Dark=FAIL)}
\label{fig:umap1}
\end{figure}
\begin{figure}[h]
\centering
\includegraphics[width=\linewidth]{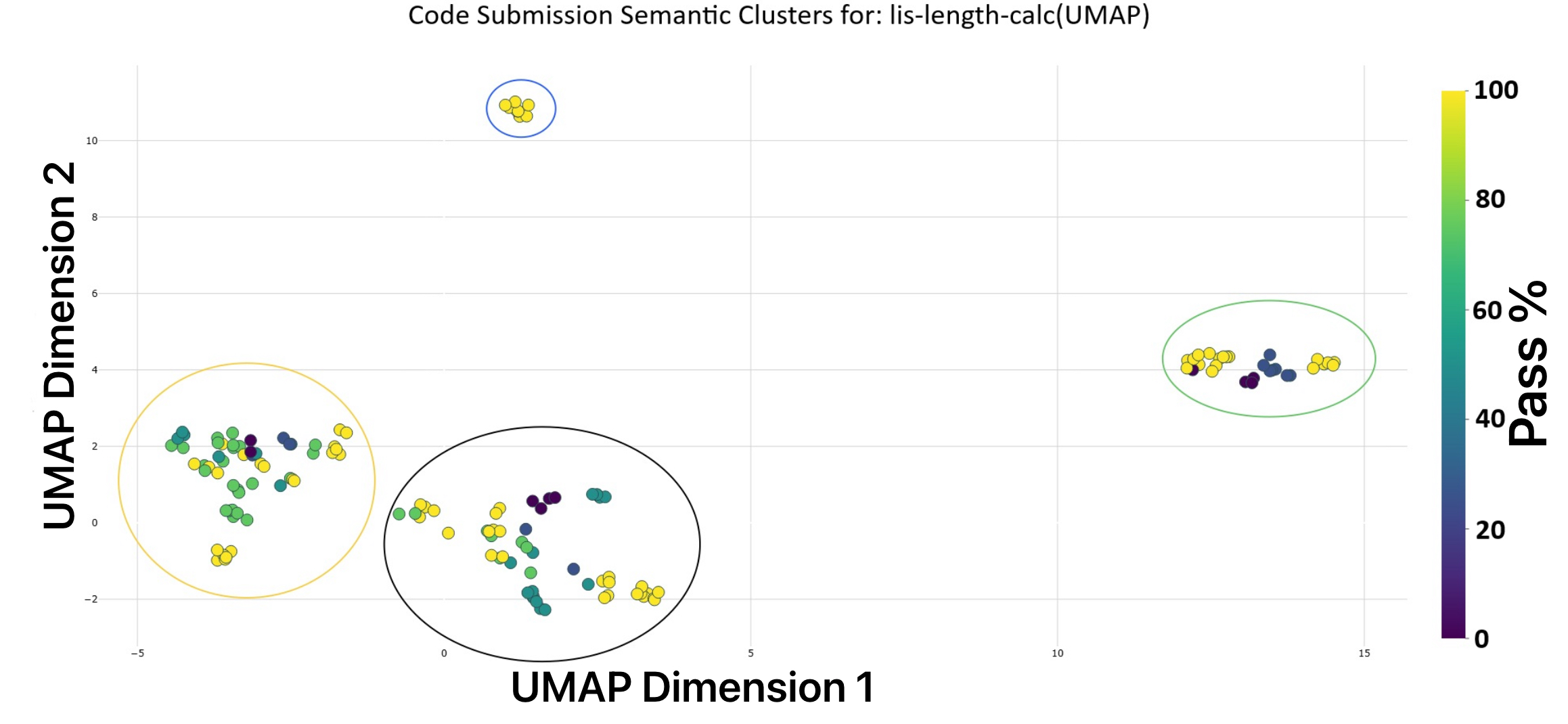}
\caption{UMAP projection of embeddings after contrastive fine tuning with the points shaded by their performance tier (e.g., Light=PASS, Dim=PARTIAL PASS, Dark=FAIL)}
\label{fig:umap2}
\end{figure}


\subsection{Qualitative Analysis: Instructor Analytics via UMAP}
A key result of our framework is its ability to generate actionable insights for instructors. Using the performance-aware embeddings produced by our contrastively trained embedding model, we generated 2D visualizations of student submissions using the Uniform Manifold Approximation and Projection (UMAP) algorithm. In Fig-5, effect of Contrastive Fine-Tuning is clearly visible as distinct clusters based upon different approaches taken to solve the question have evolved in the UMAP generated via contrasively fine-tuned embedding model. The clusters represent various distinct approaches taken to solve a particular question.

\section{Future Work}
While Autograder+ demonstrates a novel integration of AI-driven semantic feedback, prompt pooling, and performance-aware visual analytics, several avenues are open for expansion and refinement:\\
\textbf{Classroom Deployment:} Implement Autograder+ in programming courses, evaluate its practical impact on learner experience, feedback quality, and instructional workflows. This will provide the evidence of its effectiveness and reveal challenges in integration with existing teaching practices.\\
\textbf{Longitudinal Learning Analytics: }Assess it's effects on problem-solving strategies, misconceptions, and self-efficacy, using temporal UMAPs to track individual and cohort evolution with time.\\
\textbf{Large-Scale Evaluation: }Deployment across diverse institutions and track its long-term impact on learner performance, scalability, and instructor adoption.\\
\textbf{Cross Domain Generalization: }Extend adaptability beyond introductory programming to domains like systems, DSA etc.

\section{Conclusion}
We presented Autograder+, designed to address the core challenges faced by faculty in large-scale programming courses: balancing scalability with meaningful, individualized feedback. By combining domain-specific LLM fine-tuning, performance-aware semantic embeddings, and dynamic prompt pooling, the system produces feedback that diagnoses functional errors and surfaces underlying conceptual gaps in a manner aligned with instructional goals.
For instructors, Autograder+ offers actionable, visual analytics that reveal common misconceptions, alternative solution strategies, and at-risk students early in the learning process. This enables targeted interventions, reduces grading overhead, and frees faculty time for deeper engagement with students. 
Our next steps involve piloting it in our own programming courses to test its feasibility, refine its components, and collect empirical data on its effectiveness. If successful, the framework’s modular design allows
adaptation to other disciplines where structured problem-solving and conceptual mastery are central, making it a potentially sustainable, faculty-centric solution for delivering high-quality feedback at scale.

\bibliographystyle{ACM-Reference-Format}
\bibliography{sample-base}

\end{document}
\endinput